\documentclass[lettersize,journal]{IEEEtran}
\usepackage{amsmath,amsfonts}
\usepackage{algorithmic}
\usepackage{algorithm}
\usepackage{array}
\usepackage{amsmath}
\usepackage{booktabs}
\usepackage[caption=false,font=normalsize,labelfont=sf,textfont=sf]{subfig}
\usepackage{enumitem}
\usepackage{textcomp}
\usepackage{stfloats}
\usepackage{url}
\usepackage{verbatim}
\usepackage{graphicx}
\usepackage{multirow}
\usepackage{threeparttable}
\usepackage{pifont}
\usepackage{cite}
\usepackage{cuted}
\usepackage{capt-of}
\usepackage{graphicx}
\usepackage[colorlinks,bookmarksopen,bookmarksnumbered,citecolor=blue, linkcolor=blue, urlcolor=blue]{hyperref}
\usepackage{xcolor}  

\newcommand{\wrongmark}{\textcolor{red}{\ding{53}}}

\begin{document}

\title{LL-Localizer: A Life-Long Localization System based on Dynamic i-Octree}

\author{
Xinyi Li$^{1}$, Shenghai Yuan$^{2}$, Haoxin Cai$^{1}$, Shunan Lu$^{1}$, Wenhua Wang$^{1}$, Jianqi Liu$^{1}$\textsuperscript{*}

\thanks{\textsuperscript{*}Corresponding Author. This work was supported in part by the National Science Foundation of China under Grant 62172111, National Joint Fund Key Project (NSFC - Guangdong Joint Fund) under Grant U21A20478.  }
\thanks{$^{1}$Xinyi Li, Haoxin Cai, Shunan Lu, Wenhua Wang, Jianji Liu are with School of Computer Science, Guangdong University of Technology, Guangzhou 510006, China. Email:  
 xinyili@mail2.gdut.edu.cn , hxcai@mail2.gdut.edu.cn, shunanlu@mail2.gdut.edu.cn, wenhuawang@mail2.gdut.edu.cn,  liujianqi@ieee.org. \par
$^{2}$Shenghai Yuan is with Centre for Advanced Robotics Technology Innovation,  Nanyang Technological University, Singapore. Email: shyuan@ntu.edu.sg\par
}
}
\maketitle
\begin{strip}
\begin{minipage}{\textwidth}\centering
\vspace{-110pt}
\includegraphics[width=0.95\textwidth]{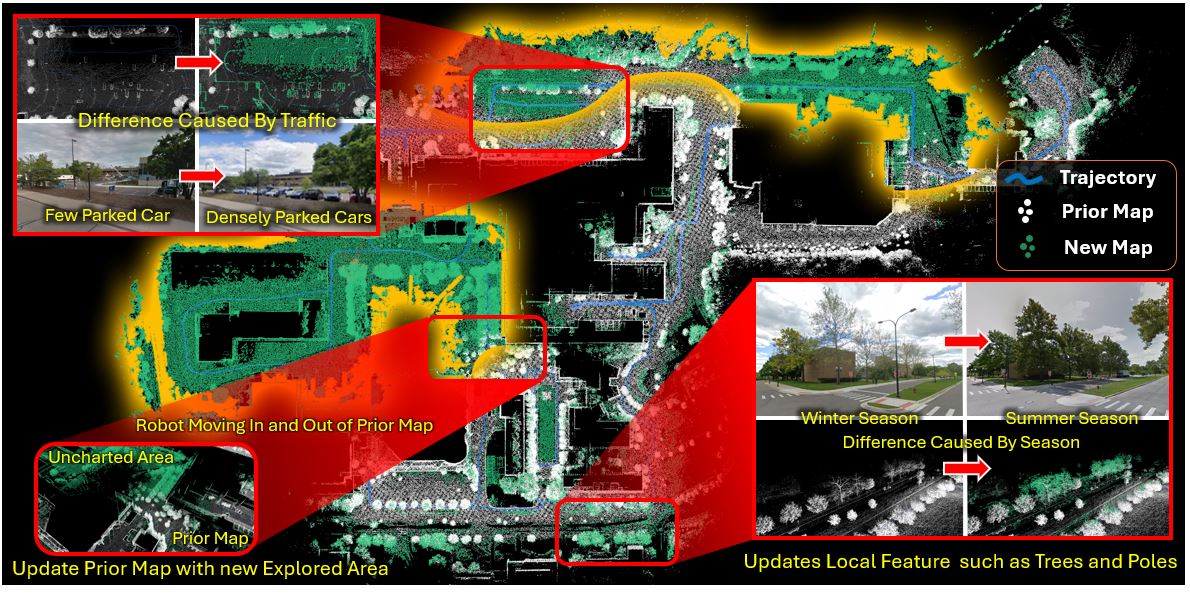}
\vspace{-10pt}
\captionof{figure}{Motivation for our incremental voxel-based life-long localization approach. The white point cloud denotes the prior map loaded at startup, while the green point cloud highlights incremental updates to accommodate new areas or environmental changes (e.g., traffic, seasons).  This functionality is crucial for long-duration service robots.}
\label{introduction}
\vspace{-15pt}
\end{minipage}
\end{strip}

\begin{abstract}
This paper proposes an incremental voxel-based life-long localization method, LL-Localizer, which enables robots to localize robustly and accurately in multi-session mode using prior maps. Meanwhile, considering that it is difficult to be aware of changes in the environment in the prior map and robots may traverse between mapped and unmapped areas during actual operation, we will update the map when needed according to the established strategies through incremental voxel map.
Besides, to ensure high performance in real-time and facilitate our map management, we utilize Dynamic i-Octree, an efficient organization of 3D points based on Dynamic Octree to load local map and update the map during the robot's operation.
The experiments show that our system can perform stable and accurate localization comparable to state-of-the-art LIO systems. And even if the environment in the prior map changes or the robots traverse between mapped and unmapped areas, our system can still maintain robust and accurate localization without any distinction. Our demo can be found on Blibili \url{https://www.bilibili.com/video/BV1faZHYCEkZ} and youtube \url{https://youtu.be/UWn7RCb9kA8} and the program will be available at \url{https://github.com/M-Evanovic/LL-Localizer}.
\end{abstract}

\begin{IEEEkeywords}
Robotics, Life-long, Localization.
\end{IEEEkeywords}

\section{Introduction}

\IEEEPARstart{L}{ocalization} is a key technology for mobile robots, enabling them to navigate autonomously in diverse environments. Life-long localization utilizes prior maps to achieve localization and map updates across multi-session. While many state-of-the-art LiDAR odometry (LO) and LiDAR-inertial odometry (LIO) systems \cite{fast-lio2, chen2023direct, faster-lio,kiss-ICP,zhang2023ri,zheng2024traj,yuan2025semielastic,zheng2024trajlio,tao2025equivariant} perform well in localization and mapping, they are designed more for single-session scenarios rather than multi-session applications. Consequently, their ability to handle spatial and temporal correlations is limited, making it challenging to extend them to life-long localization and mapping mode. \par

In life-long deployment, prior maps are commonly used for localization. However, even with high-precision prior maps, life-long localization still faces several challenges. Map-based localization methods \cite{Posemap, HDL-Loc} directly constrain the current pose by aligning the current scan with the prior map. However, if the environment in the prior map changes or the robot enters an area without prior map, pose estimation will become unreliable due to inaccurate or insufficient prior knowledge. Although the Global Navigation Satellite System (GNSS) can directly provide the robot's absolute position, it has an inherent limitation, as GNSS data may be unavailable or unreliable in GNSS-denied environments. \par

In practical life-long deployment, robots can enter unknown areas (yellow regions highlighted in Fig. \ref{introduction}) where there is no prior map during autonomous navigation. If the robot fails to recognize that it has entered an unknown area and does not update the surrounding map, localization failure may occur, leading to system failure. Likewise, if environmental changes occur within the prior map region (red box in Fig. \ref{introduction}), point cloud registration may fail, resulting in significant localization errors. \par

To address these limitations, we propose LL-Localizer, which focuses on tackling life-long localization challenges: even as the robot traverses between mapped and unmapped areas or when the environment within the map changes, it can still maintain robust localization while simultaneously updating the map. \par

The main contributions in this paper are summarized as follows:
\begin{itemize}
    \item With the proposed map management approach, the system can autonomously and flexibly switch between map-based localization and SLAM modes during operation, ensuring consistently accurate and robust localization in both mapped and unmapped areas.
    \item Our local map loading method, built upon the proposed map management approach, enables accurate point cloud registration to ensure robust localization even when environmental changes occur within the prior map. At the same time, our system allows updating the map to reflect these changes.
    \item We propose Dynamic i-Octree for efficient map management, greatly accelerating local map loading and map updates. This enhancement allows our system to maintain real-time performance even in dense maps.
\end{itemize}

\section{Related Works}
\subsection{LO / LIOs}

The Iterative Closest Point (ICP) method introduced by Besl et al. \cite{ICP1} has long served as a fundamental technique for scan registration \cite{ICP2}, forming the basis of many LiDAR Odometry (LO) systems. While ICP performs well on dense 3D scans with accurate point correspondences, its strict reliance on exact matches makes it less suitable for sparse point clouds. To address this limitation, Segal et al. proposed the generalized-ICP (G-ICP) algorithm \cite{GICP}, which leverages point-to-plane distances instead of exact matches. Building on this idea, Zhang et al. extended the model by introducing point-to-edge distances, leading to the development of the LOAM framework \cite{LOAM}. Variants such as LeGO-LOAM \cite{LeGO-LOAM} and F-LOAM \cite{F-LOAM} were further optimized for structured and unstructured environments, respectively. In parallel, surfel-based SLAM systems \cite{surfel-based-SLAM} have also emerged, with SuMa++ \cite{SuMa++} enhancing them through the integration of fully convolutional neural networks for semantic information extraction. However, these methods can struggle in low-feature environments or when using LiDARs with a narrow field of view, prompting the need for more robust solutions.

To handle such challenges, a wave of more adaptive and computationally efficient LiDAR-inertial odometry (LIO) methods has been proposed. FAST-LIO2 \cite{fast-lio2}, built upon the iKD-tree \cite{iKD-tree}, uses an iterative extended Kalman filter (iEKF) to perform real-time scan-to-map alignment, and later replaces handcrafted features with point-to-plane ICP over raw points, significantly improving generalization. Faster-LIO \cite{faster-lio} introduced sparse incremental voxels (iVox) for faster updates and efficient nearest neighbor search. Kiss-ICP \cite{kiss-ICP} relies on adaptive thresholding in a point-to-point ICP framework, combined with downsampling strategies to ensure both robustness and speed. To further tackle sensor-level challenges, Shi et al. \cite{shi2023motion} proposed a motion distortion elimination algorithm based on multioutput Gaussian process regression (MOGPR). iG-LIO \cite{iG-LIO} incorporated incremental G-ICP and voxel-based surface covariance estimation \cite{voxel_map}, maintaining accuracy while improving computational efficiency. IGE-LIO \cite{chen2024ige} expanded on this by integrating LiDAR intensity gradients, offering superior resilience in degenerate scenes. Other contributions such as Xu et al.’s intermittent VIO for degeneracy mitigation \cite{xu2024intermittent}, LVINS’s fusion of LiDAR and visual descriptors \cite{zhao2024multi}, and Doppler-based odometry with vehicle kinematics \cite{pang2024efficient}, further demonstrate the increasing depth and breadth of LIO research.

Recent methods continue to expand capabilities in dense mapping, fusion, and long-term deployment. SuIn-LIO \cite{zhang2024high} leverages invariant EKF and efficient surfel mapping to achieve high performance, while Tao et al. \cite{tao2024lidar} proposed a point-to-likelihood HD map matching method to improve tunnel localization. Gao et al. \cite{gao2024robust} demonstrated robust fusion odometry using INS-centric multi-modal systems, and Wang et al. \cite{wang2024range} developed a TSDF-based dense mapping pipeline with augmented LiDAR data. MM-LINS and BA-LINS \cite{tang2025ba} extend this progress through multi-map management and frame-to-frame bundle adjustment, respectively. SE-LIO \cite{zhang2025se} incorporates semantic cues and adaptive cylinder fitting for reliable navigation in forested areas, while TLS-SLAM \cite{cheng2025tls} employs 3D Gaussian splatting to improve convergence in large-scale mapping. ININ-LIO \cite{gao2025inin} applies deep learning to reduce inertial drift, and methods like SGT-LLC \cite{wang2025sgt} and UA-LIO \cite{wu2025ua} demonstrate advances in loop closure and uncertainty-aware odometry. Although these systems exhibit excellent localization accuracy and robustness, they remain limited to single-session operation \cite{nguyen2021viral,lou2025qlio,ji2024sgba,tao2024silvr,zhao2024adaptive,yu20242,Xu2024M,cai2025bev}, and do not yet address the broader challenges of life-long deployment and multi-session mapping in evolving environments \cite{zhu2024swarm, border2024osprey,li20243d,li2024helmetposer}.

\subsection{Multi-Session}
Recent research has increasingly focused on multi-session frameworks to overcome the limitations of single-session systems, addressing long-term mapping and localization challenges. These approaches enhance pose estimation by incorporating additional constraints and improve adaptability to dynamic environments. For instance, K. Koide et al. \cite{HDL-Loc} developed a SLAM-based relocalization framework with an independent odometry module that leverages scan-to-map information to refine pose estimates. Similarly, Giseop Kim et al. \cite{LT-mapper} introduced a modular LiDAR-based life-long mapping framework tailored for urban environments. Their multi-session SLAM approach, which includes dynamic change detection and change management, effectively manages trajectory errors while optimizing memory and computational costs by automatically segregating objects from large-scale point cloud maps.

Complementary methods further illustrate the benefits of multi-session strategies. Block-map-based localization \cite{block_loc} employs a factor graph that integrates IMU pre-integration and scan-matching factors to enhance localization accuracy, while a range-inertial localization algorithm \cite{range-inertial-loc} tightly couples scan-to-scan and scan-to-map registrations with IMU data within a sliding-window factor graph framework. LTA-OM \cite{LTAOM} advances LiDAR SLAM with innovations such as complete loop detection and correction, false-positive loop closure rejection, long-term association mapping, and multisession localization—utilizing a corrected history map to achieve drift-free odometry at revisit locations. Additionally, LiLoc \cite{LiLoc} presents a graph-based life-long localization framework that maintains a single central session and leverages multimodal factors for enhanced accuracy and timeliness. Building on these foundations, our proposed method supports life-long deployment by delivering online, accurate, and robust localization through prior maps, seamlessly integrating environmental changes and unmapped areas into the original map.

\section{Methodology}
This section covers key concepts, such as definitions for life-long map management, an improved data structure, the overall system framework, and the algorithmic implementation.

\subsection{Definition}\label{definition}
To establish a rigorous foundation, we first introduce the fundamental notations employed in this study. The state of the robot $\mathbf{x}$ is denoted as:
\begin{equation}
    \mathbf{x} = [\mathbf{R}^{\top}, \mathbf{p}^{\top}, \mathbf{v}^{\top}, \mathbf{b}^{\top}]^{\top}
\end{equation}
where $\mathbf{R} \in SO(3)$ is the rotation matrix. $\mathbf{p} \in \mathbb{R}^3$ is the position vector. $\mathbf{v} \in \mathbb{R}^3$ is the speed vector. $\mathbf{b}$ is the bias of IMU. The full robot pose is represented as $\mathbf{T} = (\mathbf{R},\mathbf{p}) \in SE(3)$. 
Moreover, $\mathbf{x}$, $\hat{\mathbf{x}}$ and $\overline{\mathbf{x}}$ denote  the ground-true, predicted and updated values of $\mathbf{x}$.
$B$ and $M$ refer to the robot's body coordinate frame and the map coordinate frame. \par

Additionally, to effectively manage map updates and localization, we define the following map components. Let the map \(\mathcal{M}\) be generated from a set of poses \(\mathbf{T}_i\) and corresponding sensor data \(\mathbf{S}_i\), where each scan \(\mathbf{S}_i = \{ \mathbf{p}_j \mid \mathbf{p}_j \in \mathbb{R}^3, j = 1, \ldots, N_i \}\) is a point cloud of \(N_i\) 3D points.

%
\begin{enumerate}[label=(\alph*)]
    \item \textbf{Prior Map Points \(\mathcal{M}^p\) and Prior Blocks \(\mathcal{B}^p\):}  
    \[
        \mathcal{M}^p = \bigcup_{i=1}^N \mathbf{T}_i \,\mathcal{S}_i,\quad
        \mathcal{B}^p = \mathcal{V}(\mathcal{M}^p),
    \]
    where \(\mathcal{V}(\mathcal{P})\) partitions a point set \(\mathcal{P}\) into its constituent voxel blocks. The prior map is the union of historical scans at their respective poses, and the voxel blocks containing \(\mathcal{M}^p\) form the prior blocks \(\mathcal{B}^p\).

    \item \textbf{Temporary Map Points \(\mathcal{M}^t\):}  
    \[
        \mathcal{M}^t = \mathbf{T}_k \,\mathcal{S}_k.
    \]
    New scan points that do not match any in \(\mathcal{M}^s\) are added to \(\mathcal{M}^t\), which serves as an auxiliary (backup) map for localization, complementing the static map. When the number of points in a voxel block exceeds a predefined threshold, these points are promoted to \(\mathcal{M}^s\).

    \item \textbf{Static Map Points \(\mathcal{M}^s\):} 
    \[
        \mathcal{M}^s = \mathcal{M}^p \cup \mathcal{M}^t.
    \]
    During operation, points from \(\mathcal{M}^t\) may be promoted and points from \(\mathcal{M}^p\) removed according to an update policy. \(\mathcal{M}^s\) is the primary reference for localization and serves as the final output map.
\end{enumerate}

\subsection{Dynamic i-Octree}
\begin{figure}
\centering
\includegraphics[width=\linewidth]{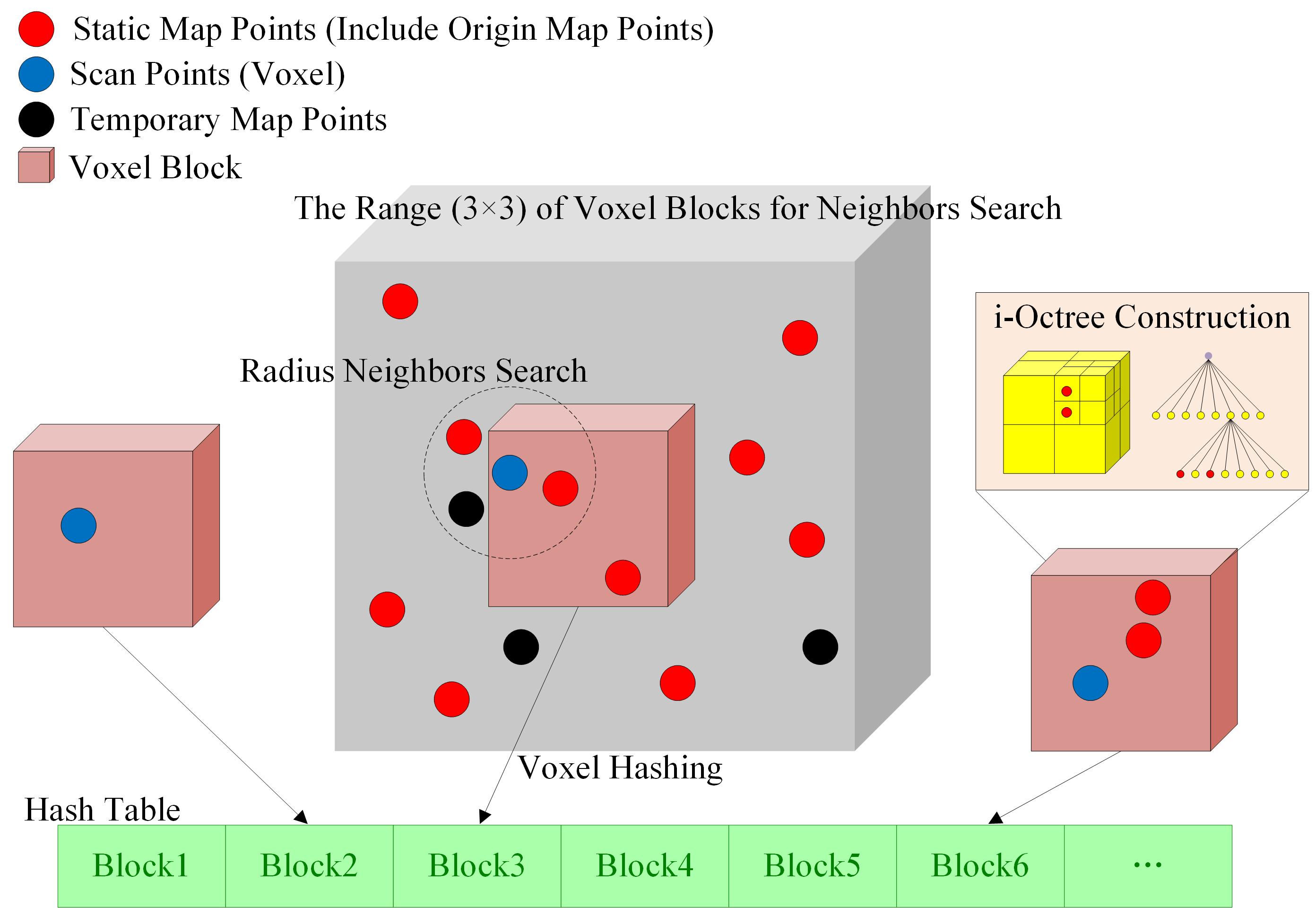}
\caption{Diagram of Dynamic i-Octree.}
\label{voxel_map}
\end{figure}

Efficiently managing these map points requires a robust structure. While Dynamic Octree \cite{Dynamic-Octree} is effective for map management, it cannot determine when to update the map or which areas need updating, limiting its ability to switch between localization and map updates.
To address this, we propose Dynamic i-Octree  (see Fig. \ref{voxel_map}), a refined version of Dynamic Octree that efficiently handles randomly distributed points. By organizing the map based on the defined structure, Dynamic i-Octree enables dynamic local map loading, improving localization robustness and facilitating map updates. For better efficiency, we leverage i-Octree \cite{i-Octree} to manage the points.

\subsubsection{Hash Pair (Key, Value)}
\begin{itemize}
    \item Voxel (Key): $\mathbf{p}$ of the point $p \in \mathbb{R}^3$ in $M$.
    \item Voxel Block (Value): the data contained in each voxel block includes $\mathcal{M}^p$, $\mathcal{M}^t$, $\mathcal{M}^s$ and an i-Octree organized by $\mathcal{M}^s$ for radius neighbor search.
\end{itemize}

\subsubsection{Hash Function Construction} \label{hash index}
Define the voxel block index vector as:
\begin{equation}
        \mathcal{B}
            =
                \begin{bmatrix}
                \mathcal{B}_x \\ \mathcal{B}_y \\ \mathcal{B}_z 
                \end{bmatrix}
            = 
                \begin{bmatrix}
                \lfloor \frac{p_x}{s} \rfloor \\ 
                \lfloor \frac{p_y}{s} \rfloor \\ 
                \lfloor \frac{p_z}{s} \rfloor 
                \end{bmatrix} \\
\end{equation}
where $p=[p_x, p_y, p_z]^{\top} \in \mathbb{R}^3$ is the 3D coordinate of the point in $M$. The parameter $s$ is the resolution of each voxel block. $\lfloor \cdot \rfloor$ denotes the floor function, which rounds down to the nearest integer. $\mathcal{B}_x$, $\mathcal{B}_y$, $\mathcal{B}_z$ are voxel block indices.
The hash index for the voxel block is computed as:
\begin{equation}
    \mathcal{H}_{\mathcal{B}} = (\mathcal{B}_x\eta_x)\oplus(\mathcal{B}_y\eta_y)\oplus(\mathcal{B}_z\eta_z)
\end{equation}
where $\mathcal{H}_{\mathcal{B}}$ represents the computed hash valus. $\eta{_x}$, $\eta{_y}$ and $\eta{_z}$ are large prime numbers(e.g., $\eta{_x}=73856093$, $\eta{_y}=19349669$, $\eta{_z}=83492791$) to reduce collisions and ensure good hashing properties. $\oplus$ denotes the bitwise XOR operator.

\subsubsection{i-Octree}
 In our system, we will retain the prior map as possible to ensure the accuracy of localization and map integrity, which will lead to significant differences in the density of points within the voxel blocks, with some voxel blocks having particularly high point densities. In this case, using i-Octree to organize $\mathcal{M}^s$ for radius neighbor search instead of traversal can greatly improve the efficiency. \par
 
\begin{enumerate}[label=(\alph*)]
    \item Implementation of i-Octree: we implement i-Octree in the voxel block, but we have made some modifications to i-Octree to make it more suitable for our system. Because we implement i-Octree in a voxel block of a certain size, the boundary of the axis-aligned bounding box of i-Octree is determined, we don't need to consider the situation that any points may be beyond the boundary. So we don't need to create new root octant to expand the bounding box. Besides, in order to preserve more of the prior map and maintain robustness in locating in areas without map, we have to sacrifice some efficiency to improve the accuracy of localization. Therefore, we only retained the downsampling function when inserting new points, but removed the Box-wise Delete function. \par
    \item Radius Neighbor Search: for each query point $q \in \mathbb{R}^3$ and the radius of neighbor search $r$, radius neighbor search finds every point $p$ satisfying $||p-q||_2<r$. To accelerate radius neighbor search, i-Octree adopt the pruning strategy proposed by Behley et al. \cite{Octree} with improvements to reduce computation cost. In addition, we further improve efficiency by parallel computing the radius neighbor search of each point in the scan. \par
\end{enumerate}

\subsection{System Framework}
\begin{figure*}[!t]
\centering
\includegraphics[width=\linewidth]{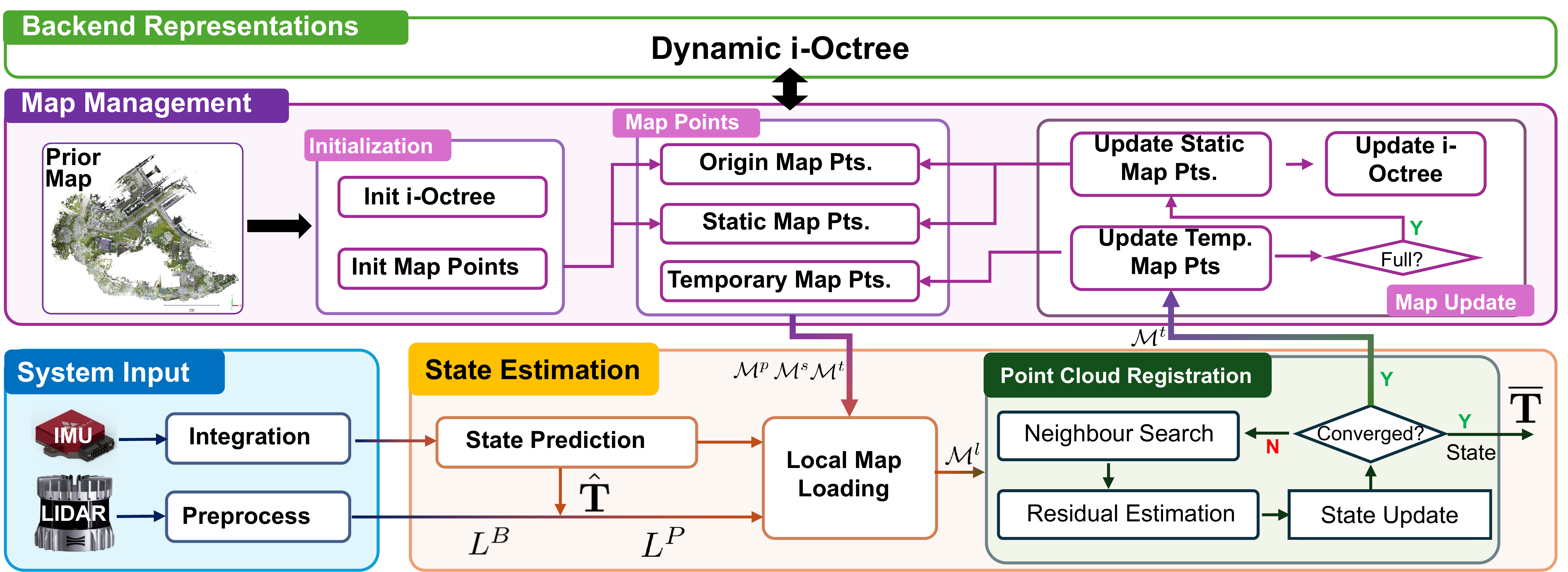}
\vspace{-15pt}
\caption{System overview of LL-Localizer.}
\label{overview}
\vspace{-15pt}
\end{figure*}

The overview of our proposed framework is shown in Fig. \ref{overview}.
Before running the robot, we load a prior map first. Then each point in this map will be added to $\mathcal{M}^p$ and $\mathcal{M}^s$ in the voxel block which it belongs to, and i-Octree will be constructed according to $\mathcal{M}^s$. \par

When the first few scans arrive, we merge them and then obtain the initial pose $T_0$ within the map using the global localization module Branch-and-Bound Search++ \cite{block_loc} (BBS++).
During system operation, when a new LiDAR scan $L^B$ arrives, the system transforms $L^B$ according to the predicted pose $\hat{\mathbf{T}}$ to obtain $L^P$. It then dynamically loads the local map $\mathcal{M}^l$ following the strategy (see Section \ref{strategy}), and performs point cloud registration to obtain accurate pose $\overline{\mathbf{T}}$ in $M$. After these processes, points in $L^P$ that do not match $\mathcal{M}^s$ will be added to the $\mathcal{M}^t$ for future map updates. Once the size of $\mathcal{M}^t$ in a voxel block reaches the predefined threshold, $\mathcal{M}^s$ will be updated, and i-Octree will be updated accordingly (see Section \ref{Map Update}). 
After the operation ends, $\mathcal{M}^s$ will be output as the final result map.

\subsection{IMU Integration}
Through the high-frequency input of IMU data, the system can calculate the pose of the robot with more continuity.\par

The measurements of acceleration $\hat{a}$ and angular velocity $\hat{\omega}$ from IMU at time $t$ are defined as follows:
\begin{align}
    \hat{\alpha}_t &= \alpha_t-g+\mathbf{b}_t^\alpha+n_t^\alpha \\
    \hat{\omega}_t &= \omega_t+\mathbf{b}_t^\omega+n_t^\omega
\end{align}
where $\hat{\alpha}_t$ and $\hat{\omega}_t$ denote the raw IMU measurements in $B$ at $t$. The $\alpha_t$ and $\omega_t$ are the ground-truth values in $B$ at $t$. $g$ is the gravity vector in map frame $M$. The $\mathbf{b}$ and $n$ represent bias and white noise of IMU, respectively.\par

Then, the system can predict the state including the pose $\mathbf{T}^P$ of the robot from time $t$ to time $t+\Delta t$ until the next LiDAR scan arrives as follows:
\begin{align}
    \hat{\mathbf{v}}_{t+\Delta t} &= \hat{\mathbf{v}}_t + g \Delta t + \hat{\mathbf{R}}^{MB}_t(\hat{\alpha}_t - \mathbf{b}_t^\alpha - n_t^\alpha) \Delta t \\
    \hat{\mathbf{p}}_{t+\Delta t} &= \hat{\mathbf{p}}_t + \hat{\mathbf{v}}_t \Delta t + \frac{1}{2} g \Delta t^2 + \frac{1}{2} \hat{\mathbf{R}}^{MB}_t ( \hat{\alpha}_t - \mathbf{b}_t^a - n_t^\alpha ) \Delta t^2 \\
    \hat{\mathbf{R}}_{t+\Delta t}^{MB} &= \hat{\mathbf{R}}_t^{MB} Exp((\hat{\omega}_t - \mathbf{b}_t^\omega - n_t^\omega) \Delta t)
\end{align}
where $Exp(\omega) = exp(\omega^\wedge)$, and $exp(\cdot)$ is the exponential map function in SO(3). $\hat{\mathbf{R}}_t^{MB}$ is the rotation matrix from baseframe $B$ to map $M$ at $t$. Based on the above calculation, the system can make a preliminary estimation of the robot's state.

\subsection{Scan Process}
Predicting the robot's pose solely based on IMU is not accurate enough, especially after long-duration and long-distance movements, where significant errors may accumulate. Therefore, we need to process LiDAR scans to correct the predicted pose $\mathbf{T}^P$.
When a new scan arrives, system processes it through the following series of steps to achieve accurate localization and determine whether the map needs to be updated simultaneously.

\subsubsection{Local Map $\mathcal{M}^l$ Loading Strategy}\label{strategy}
With each incoming scan, the system loads the appropriate local map $\mathcal{M}^l$ based on the scanned points. System first transform the point cloud from body coordinate to world coordinate through IMU integration. The system then locates the corresponding voxel block using the hash function in \ref{hash index} and loads the appropriate map points from the three types of maps points to improve localization robustness. The loading strategy is mainly based on two criteria: the voxel block $\mathcal{B}^r$ where the robot is located and the number \textit{num} of scanned points in $\mathcal{B}^p$.\par

To evaluate the distribution of the scanned points, we first define the Heaviside Step function $H(x)$ as:
\begin{equation} 
H(B^r, B^p) = 
\begin{cases} 
0, & \mathcal{B}^r \in \mathcal{B}^p, \\ 
1, & \mathcal{B}^r \notin \mathcal{B}^p. 
\end{cases} 
\end{equation}

Then, the effective distance $\phi^*$ can be defined as:
\begin{equation} 
\phi^{*} = \phi_2 + (\phi_1 - \phi_2) H(B^r, B^p), 
\end{equation}
where $\phi_1$ and $\phi_2$ are predetermined factors, with $\phi_1 > \phi_2$. These factors can be adjusted to control the map loading strategy. They are initially chosen heuristically now and can be further optimized through a reinforcement learning framework to adapt to different environments and scanning conditions.

Next, we compute the ratio $\rho$ to evaluate the spatial distribution of scanned points. For a LiDAR with a narrow field of view, we define $\rho_n$ as:  
\begin{equation}
\rho_n =
\frac{\phi^{*2} \tan{\frac{\theta_L}{2}}}{\frac{1}{2} \times \phi_e \times \theta_L} \Big( 1 - 2 H(B^r, B^p) \Big) + H(B^r, B^p)
\end{equation}
where:  
\begin{itemize}
    \item $\phi^*$ is the effective distance, as defined earlier.
    \item $\phi_e = d_{max}^2 - d_{min}^2$ is the effective range of the LiDAR.
    \item $\theta_L$ is the horizontal field of view of the LiDAR, (in Rad).
\end{itemize}

For a LiDAR with a $360^{\circ}$ horizontal field of view, the ratio $\rho_w$ is defined as:  
\begin{equation}
\rho_w = \frac{1}{\pi d_{max}^2} \int_{-\sqrt{d_{max}^2 - \phi^{*2}}}^{\sqrt{d_{max}^2 - \phi^{*2}}} (\sqrt{d_{max}^2 - \mathbf{x}^2} - \phi^*) d\mathbf{x}
\end{equation}

Finally, the threshold $\tau$ is computed as:  
\begin{equation}
\tau = \lfloor n \times \rho \times s \rfloor
\end{equation}
where:  
\begin{itemize}
    \item $n$ is the number of scanned points.
    \item $s$ is the resolution of each voxel block.
    \item $\rho$ is the ratio selected from $\rho_w$ or $\rho_n$ based on the actual hardware.
    \item $\lfloor \cdot \rfloor$ denotes the floor function that rounds down to the nearest integer.
\end{itemize}

And we difine a function $S(\cdot)$ to evaluate the convergence of point cloud registration:
\begin{equation} 
    S(r_k, k) = 
    \begin{cases} 
    1, & r_k > \delta\ \& \ k \leq 10, \\ 
    0, & r_k > \delta\ \& \ k > 10. 
    \end{cases} 
\end{equation}
where $r_k$ denotes the residual at iteration $k$. $\delta$ is the predefined convergence threshold. \par

After computing the threshold $\tau$ and function $S(\cdot)$, the system determines the appropriate map loading strategy based on the voxel block $\mathcal{B}^r$ where the robot is located, the number of scanned points $\kappa$ in $\mathcal{B}^p$ and point cloud registration convergence evaluation. Depending on these factors, the system can handle different localization scenarios as follows:
\begin{enumerate}[label=(\alph*)]
    \item $\mathcal{B}^r \in \mathcal{B}^p$ and $\kappa \geq \tau$:
    We consider this to be the most ideal situation: the environment in prior map has not changed, or the subtle changes in the map can be ignored. In this case, system can only load $\mathcal{M}^p$ in $\mathcal{B}^p$ as the $\mathcal{M}^l$:
    \[
        \mathcal{M}^l \subseteq \mathcal{M}^p
    \]
    which can improve the efficiency of the system while achieving high robustness localization.
    \item $\mathcal{B}^r \in \mathcal{B}^p$ and $\kappa < \tau$:
    We consider that the environment may have changed, or the robot may be moving out of the pre-mapped area into a non-mapped area. In this case, system will load $\mathcal{M}^p$ in $\mathcal{B}^p$ and $\mathcal{M}^s$ in non $\mathcal{B}^p$ as the $\mathcal{M}^l$:
    \[
        \mathcal{M}^l \subseteq (\mathcal{M}^p \cup \mathcal{M}^s)
    \]
    To ensure the robustness of localization, the system will trust the points in $\mathcal{B}^p$ more, so we assign greater weight $w^g$ to $\mathcal{M}^p$ in $\mathcal{B}^p$ in the residual calculation.calculation.

    \item $\mathcal{B}^r \notin \mathcal{B}^p$ and $\kappa \geq \tau$:  
    This occurs when the robot transitions from an unmapped area to a pre-mapped region. The system can directly load $\mathcal{M}^p$ in $\mathcal{B}^p$ as the $\mathcal{M}^l$ to quickly eliminate positional drift accumulated during localization in unmapped area. However, if the drift error is too large, point cloud registration may fail to converge. In such cases, the system incorporates additional map data $\mathcal{M}^s$ from outside $\mathcal{B}^p$ to gradually eliminate the drift error:  
    \[
    \mathcal{M}^l \subseteq 
    \begin{cases}
    \mathcal{M}^p,                    & S(r_k, k) = 1, \\
    \mathcal{M}^p \cup \mathcal{M}^s, & S(r_k, k) = 0.
    \end{cases}
    \]
    Similar to the previous case, we assign greater weight $w^g$ to $\mathcal{M}^p$ in $\mathcal{B}^p$ during residual calculation.
    
    \item $\mathcal{B}^r \notin \mathcal{B}^p$ and $\kappa < \tau$:  
    This represents an extreme case where the robot may be navigating a new environment or the pre-mapped area has undergone significant changes. In this case, the system initially loads $\mathcal{M}^s$ as the $\mathcal{M}^l$. If the information in $\mathcal{M}^s$ is insufficient for point cloud registration to converge, the system incorporates additional map data $\mathcal{M}^t$ into the $\mathcal{M}^l$:  
    \[
    \mathcal{M}^l \subseteq 
    \begin{cases}
    \mathcal{M}^s,                    & S(r_k, k) = 1, \\ 
    \mathcal{M}^s \cup \mathcal{M}^t, & S(r_k, k) = 0.
    \end{cases}
    \]
    In the residual calculation, we assign a lower weight \( w^l \) to $\mathcal{M}^t$ to reduce the impact of outliers.

\end{enumerate}

\begin{algorithm}[!h]
    \caption{Local Map Loading}
    \label{algorithm}
    \renewcommand{\algorithmicrequire}{\textbf{Input:}}
    \renewcommand{\algorithmicensure}{\textbf{Output:}}
    \begin{algorithmic}[1]
        \REQUIRE $\mathcal{M}^p$, $\mathcal{M}^s$, $\mathcal{M}^t$
        \ENSURE $\mathcal{M}^l$, weights for residual calculation $w_n$, $w_g$, $w_l$
        \STATE Translate robot's $\mathbf{T}$ from $B$ to $M$
        \IF {$\mathcal{B}^r \in \mathcal{B}^p$ and $\kappa \geq \tau$} 
            \STATE $\mathcal{M}^l$ $\leftarrow$ $\mathcal{M}^p$ $\in$ $\mathcal{B}^p$
            \STATE $\mathcal{M}^p$ $\leftarrow$ $w^n$
        \ELSIF{$\mathcal{B}^r \in \mathcal{B}^p$ and $\kappa < \tau$}
            \STATE $\mathcal{M}^l$ $\leftarrow$ $\mathcal{M}^p$ $\in$ $\mathcal{B}^p$, $\mathcal{M}^l$ $\leftarrow$ $\mathcal{M}^s$ $\notin$ $\mathcal{B}^p$
            \STATE $\mathcal{M}^p$ $\leftarrow$ $w^g$, $\mathcal{M}^s$ $\leftarrow$ $w^n$
        \ELSIF{$\mathcal{B}^r \notin \mathcal{B}^p$ and $\kappa \geq \tau$}
            \STATE $\mathcal{M}^l$ $\leftarrow$ $\mathcal{M}^p$ in $\mathcal{B}^p$
            \STATE $\mathcal{M}^p$ $\leftarrow$ $w^n$
            \IF{$S(r_k, k) = 0$}
                \STATE $\mathcal{M}^l$ $\leftarrow$ $\mathcal{M}^p$ $\in$ $\mathcal{B}^p$, $\mathcal{M}^l$ $\leftarrow$ $\mathcal{M}^s$ $\notin$ $\mathcal{B}^p$
                \STATE $\mathcal{M}^p$ $\leftarrow$ $w^g$, $\mathcal{M}^s$ $\leftarrow$ $w^n$
            \ENDIF
        \ELSIF{$\mathcal{B}^r \notin \mathcal{B}^p$ and $\kappa < \tau$}
            \STATE $\mathcal{M}^l$ $\leftarrow$ $\mathcal{M}^s$
            \STATE $\mathcal{M}^s$ $\leftarrow$ $w^n$
            \IF{$S(r_k, k) = 0$}
                \STATE $\mathcal{M}^l$ $\leftarrow$ $\mathcal{M}^s$, $\mathcal{M}^l$ $\leftarrow$ $\mathcal{M}^t$
                \STATE $\mathcal{M}^s$ $\leftarrow$ $w^n$, $\mathcal{M}^t$ $\leftarrow$ $w^l$
            \ENDIF
        \ENDIF
    \end{algorithmic}
\end{algorithm}

\subsubsection{Point Matching and Point Cloud Registration}
After local map $\mathcal{M}^l$ loading according to the strategy and radius neighbor search for the k-th scan, points ${p_1, p_2, ..., p_n} \subseteq L_k^P$ obtain their corresponding neighbor points $Q_1, Q_2, ..., Q_n \subset Q$. These neighbor points are then used to fit lines or planes. For a point $p^p \in (\mathcal{M}^l \cap Q)$ on the plane, its relationship with the plane parameters is given by:
\begin{equation}
    \mathbf{n}^{\top} p^p + d = 0
\end{equation}
where $\mathbf{n}$ is the normal vector of the plane with $|\mathbf{n}|=1$, and the superscript $^\top$ denotes the transpose. $d$ is the intercept. For a point $p^l \in (\mathcal{M}^l \cap Q)$ on the line, the equation of the line $L$ can be expressed as:
\begin{equation}
    L = \mathbf{d} x + p^l
\end{equation}
where $\mathbf{d}$ is the direction vector with $|\mathbf{d}|=1$, and $x$ is the scalar parameter used to determine the position of a point on the line.
Then we can formulate the error function between the i-th query point $p_i \in L_k^P$ and the fitted plane or line of its neighboring points:
\begin{align}
    r_i^p &= w(\mathbf{n}^{\top} ( \mathbf{R} p_i + \mathbf{p}) + d ) \\
    r_i^l &= w(\mathbf{d} \times ( \mathbf{R} p_i + \mathbf{p} - p^l ))
\end{align}
where $r_i^p$ and $r_i^l$ represent the residual from $p_i$ to the fitted plane or line, respectively. $w$ is the weight assigned according to \ref{strategy} mentioned. Finally, the optimal $\mathbf{R}^{MP}$ and $\mathbf{p}^{MP}$ are obtained by iteratively minimizing the residual $r$ until convergence:
\begin{equation}
 r = \sum_{n} \| r_n^p \|^2 + \sum_{m} \| r_m^l \|^2 < \delta
\end{equation}
when $\delta$ is the predefined convergence threshold.

\subsubsection{Map Update}\label{Map Update}
Unlike most LO/LIO systems, we do not directly update scanned points to the map. Due to the high accuracy of the prior map, we tend to trust prior maps more to prevent accumulated drift errors and reject outliers. We will only update the map when there are changes in the environment in prior map or when the robot is exploring an unknown environment.\par

If there are any points found no matching $\mathcal{M}^s$ but the queried voxel blocks already exist, that means the voxel blocks which these points belong to may have changed. Or there is another situation that the robot scans an area outside the prior map so the queried voxel blocks do not exist. Both of these situations may have disastrous effects in current and future localization. Therefore, the map needs to be updated. But in our system, these points will be added to $\mathcal{M}^t$ for temporary storage first. We do not directly add these points to the global map (i.e., $\mathcal{M}^s$) for the reason that they may be outliers, and we must reduce the effects of outliers in our system and trust the origin prior map more.\par

With the operation of the system, $\mathcal{M}^t$ in each voxel block will accumulate incrementally. Once the size of $\mathcal{M}^t$ in a voxel block reaches the predefined threshold, we assume that the environment in this voxel block changes or robot scans an area outside the prior map. Then system will add these $\mathcal{M}^t$ to $\mathcal{M}^s$ to update the map. After that, system will compare the size of $\mathcal{M}^s$ and the size of $\mathcal{M}^p$ in this updated voxel block if this voxel block is $\mathcal{B}^p$. If the size of $\mathcal{M}^s$ is greater than twice the size of $\mathcal{M}^p$, also means that the size of $\mathcal{M}^t$ is greater than the size of $\mathcal{M}^p$, we assert the environment in this voxel block in prior map undergoes drastic changes and the prior map is unreliable. Then system will clear the $\mathcal{M}^s$ in this voxel block and add $\mathcal{M}^t$ to $\mathcal{M}^s$ again, and then set this $\mathcal{B}^p$ as a non $\mathcal{B}^p$. After updating the voxel block, the system will clear the $\mathcal{M}^t$ and update i-Octree according to new $\mathcal{M}^s$.

\section{Experiments} 

\subsection{Datasets, Prior Map Reconstructions, and Baselines}  
\textbf{Datasets:} The proposed system is evaluated using three widely used datasets: the North Campus Long-Term dataset (NCLT) \cite{NCLT}, M2DGR \cite{M2DGR}, and the Botanic Garden dataset (BG) \cite{BG}. Each dataset contains multiple sequences recorded at the same location under varying environmental conditions, including different scene setups, dynamic objects, and structural changes. Table \ref{environment1} summarizes the recording dates, environments, and trajectory distances for each dataset. We use the EVO \cite{evo} trajectory evaluation tool to verify the system's precision and robustness.

\vspace{-10pt}
\begin{table}[htbp]
\centering
\caption{Sequences and Environments}
\label{environment1}
\setlength{\tabcolsep}{2pt}
\renewcommand{\arraystretch}{1.3}
    \begin{tabular}{cccc}
    \hline
    Seq. & Date & Environment & Dist.(kilometer) \\
    \hline
    nclt\_1     & 2012.4.29 & long-term, dynamic objects & 3.186 \\
    nclt\_2     & 2012.6.15 & long-term, dynamic objects &  4.106 \\
    nclt\_3     & 2013.1.10 & long-term, dynamic objects & 1.139 \\
    nclt\_4     & 2012.5.11 & long-term, dynamic objects & 6.120 \\
    gate\_1    & 2021.8.4 & open square, dynamic objects & 0.248 \\
    gate\_2    & 2021.7.31 & open square, dynamic objects & 0.139 \\
    gate\_3    & 2021.7.31 & open square, dynamic objects & 0.289 \\
    street\_1   & 2021.8.6 & open square, dynamic objects & 0.340 \\
    street\_2   & 2021.8.6 & open square, dynamic objects & 0.752 \\
    street\_3   & 2021.8.6 & open square, dynamic objects & 0.423 \\
    bg\_1       & 2022.10.8 & unstructured tree, platform vibrations & 0.855 \\
    bg\_2       & 2022.10.5 & unstructured tree, platform vibrations & 0.566 \\
    bg\_3       & 2022.10.6 & unstructured tree, platform vibrations & 0.686 \\
    \hline
    \end{tabular} 
\end{table}
\vspace{-10pt}

\textbf{Prior Map Reconstructions:} To assess the system's ability to handle long-term environmental changes, we construct the prior map using the first sequence (\textbf{sequence\_1}) from each dataset. The remaining sequences simulate life-long deployment scenarios, where the system is tested under evolving environmental conditions. This setup allows us to evaluate how well the system adapts to changing environments and maintains accurate localization.  

\textbf{Comparison Selections:} 
We surveyed the literature for existing LIO methods. Many prior works either do not provide open-source implementations \cite{nguyen2021viral,shi2023motion, zhang2024high, tao2024lidar, gao2024robust, tang2025ba, zhang2025se,  cheng2025tls,gao2025inin,border2024osprey,li2024pss,li2024hcto}, or their codebases are no longer maintained and cannot be successfully built \cite{pang2024efficient, wang2024range, wu2025ua}. Other methods \cite{xu2025airslam,jin2024robust,li2025limo,xu2024selective,yuan2024large} are less relevant to our setting, as they require additional sensory inputs or networks that are not directly compatible with our system \cite{ji2022robust, yin2023segregator,li20243d,zhu2024swarm, zheng2024fast, chen2024ige, xu2024intermittent, yin2024outram, zhao2024multi, wang2025sgt,zhao2024adaptive,Xu2024M}. Ultimately, we selected Fast-LIO2 \cite{fast-lio2}, Fast-LIO-SLAM (Fast-LIO2 with Scan Context based loop closure \cite{SC}), Faster-LIO \cite{faster-lio}, F-LOAM \cite{wang2021f} and iG-LIO \cite{iG-LIO} as our baseline, given their open-source implementation and support for loop closure or incremental registration, which makes they well-suited for large-scale environments. Additionally, we compare our system against four map-based localization systems, including HDL-Loc \cite{HDL-Loc}, Fast-LIO-Loc (map-based localization method extended from Fast-LIO2), LT-AOM \cite{LTAOM}, and LiLoc \cite{LiLoc}. Among these, the map-based localization systems and our proposed system perform localization using a prior map, whereas the SLAM systems perform simultaneous localization and mapping( with loop closure).

\subsection{Life-long Deployment}
\begin{table*}[htbp]
\centering
\caption{Absolute Pose Error in terms of RMSE and Max error (meter). The best results are in \textbf{bold}.}
\vspace{-10pt}
\label{APE}
\setlength{\tabcolsep}{3pt}
\renewcommand{\arraystretch}{1.3}
\begin{threeparttable}
    \begin{tabular}{ccccccccccccccccc}
    \hline
    \multirow{2}{*}{Seq.} & \multicolumn{2}{c}{nclt\_2} & \multicolumn{2}{c}{nclt\_3} & \multicolumn{2}{c}{gate\_2} & \multicolumn{2}{c}{gate\_3} & \multicolumn{2}{c}{street\_2} & \multicolumn{2}{c}{street\_3} & \multicolumn{2}{c}{bg\_2} & \multicolumn{2}{c}{bg\_3} \\
    \cmidrule(r){2-3} \cmidrule(r){4-5} \cmidrule(r){6-7} \cmidrule(r){8-9} \cmidrule(r){10-11} \cmidrule(r){12-13} \cmidrule(r){14-15} \cmidrule(r){16-17}
     & RMSE & Max & RMSE & Max & RMSE & Max & RMSE & Max & RMSE & Max & RMSE & Max & RMSE & Max & RMSE & Max \\
    \hline
    F-LOAM \cite{wang2021f} &
    \wrongmark & \wrongmark & \wrongmark & \wrongmark & \wrongmark & \wrongmark & \wrongmark & \wrongmark & \wrongmark & \wrongmark & \wrongmark & \wrongmark & \wrongmark & \wrongmark & \wrongmark & \wrongmark \\
    Hdl-Loc \cite{HDL-Loc} &
    \wrongmark & \wrongmark & \wrongmark & \wrongmark & 0.168 & 1.159 & 0.328 & 1.028 & \wrongmark & \wrongmark & 0.176 & 0.423 & \wrongmark & \wrongmark & \wrongmark & \wrongmark \\
    Fast-LIO-Loc &
    1.988 & 3.990 & 0.911 & \textbf{2.060} & 0.153 & 0.359 & 0.313 & 0.965 & 0.299 & 0.688 & 0.183 & 0.449 & 0.523 & 1.477 & 0.358 & 0.896 \\
    LTA-OM \cite{LTAOM} &
    2.185 & 6.156 & 1.648 & 5.290 & 0.222 & 0.598 & 0.414 & 0.977 & 0.328 & 0.722 & 0.183 & 0.465 & 0.467 & 1.051 & 0.636 & 1.507 \\
    LiLoc \cite{LiLoc} &
    \wrongmark & \wrongmark & \wrongmark & \wrongmark & 0.138 & 0.315 & 0.366 & 1.005 & 0.480 & 0.964 & 0.129 & 0.261 & 0.465 & 1.164 & 0.585 & 1.474 \\
    Fast-LIO2 \cite{fast-lio2} &
    2.006 & 3.887 & 0.912 & 2.061 & 0.154 & 0.349 & 0.322 & 0.975 & 0.299 & 0.676 & 0.149 & 0.503 & 0.672 & 1.491 & 0.350 & 0.847 \\
    Fast-LIO-SLAM &
    \wrongmark & \wrongmark & \wrongmark & \wrongmark & 0.545 & 0.944 & 0.341 & 0.952 & 5.102 & 9.263 & 0.205 & 0.578 & 0.844 & 2.298 & 0.729 & 3.375 \\
    Faster-LIO \cite{faster-lio} &
    1.614 & \textbf{3.477} & 0.941 & 2.291 & 1.337 & 1.716 & 1.268 & 1.976 & 0.825 & 1.631 & 1.175 & 1.703 & 1.331 & 1.829 & 1.047 & 1.442 \\
    iG-LIO \cite{iG-LIO} &
    1.565 & 4.032 & 1.148 & 2.619 & \textbf{0.117} & \textbf{0.291} & 0.322 & 0.996 & 0.287 & 0.664 & \textbf{0.126} & \textbf{0.238} & \wrongmark & \wrongmark & \wrongmark & \wrongmark \\
    LL-Localizer (Ours) &
    \textbf{1.529} & 3.989 & \textbf{0.888} & 2.303 & 0.164 & 0.395 & \textbf{0.312} & \textbf{0.807} & \textbf{0.283} & \textbf{0.633} & 0.152 & 0.526 & \textbf{0.255} & \textbf{0.642} & \textbf{0.239} & \textbf{0.554} \\
    \hline
    \end{tabular}
    \begin{tablenotes}
        \item[1] “\wrongmark” indicates system failure due to unrecoverable drift (>10\% distance error) or numerical instability.
        \item[2] Fast-LIO-Loc is a map-based localization method adapted from Fast-LIO2 in our deployment.
        \item[3] Fast-LIO-SLAM is our own implementation of Fast-LIO2 with Scan Context.
    \end{tablenotes}
\end{threeparttable}
\end{table*}

\begin{figure*}[!t]
\centering
\includegraphics[width=\linewidth]{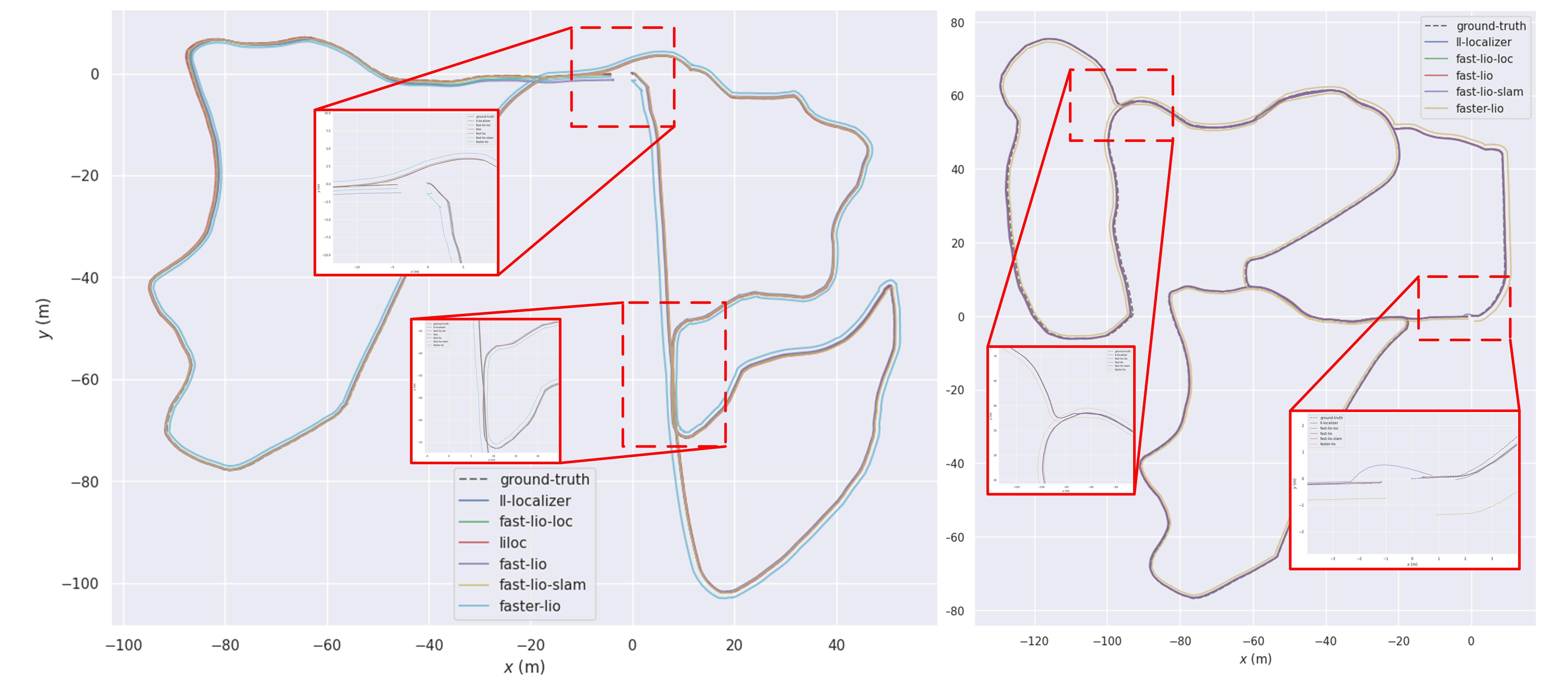}
\vspace{-25pt}
\caption{The trajectory results of different systems in bg\_2 (left) and bg\_3 (right).}
\vspace{-10pt}
\label{bg-traj}
\end{figure*}

\begin{figure*}[!t]
\centering
\includegraphics[width=\linewidth]{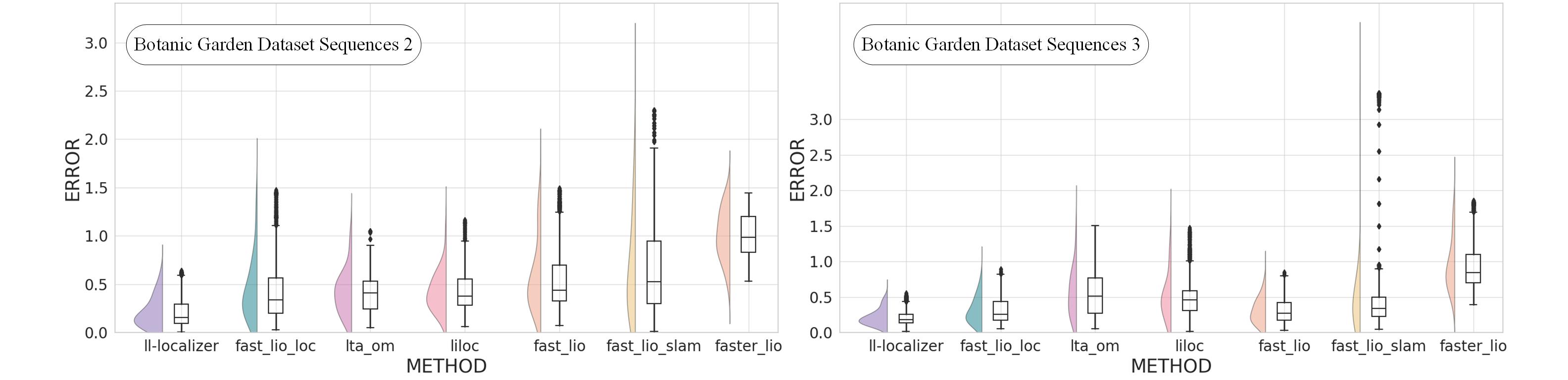}
\vspace{-25pt}
\caption{The error of different systems in bg\_2 (left) and bg\_3 (right).}
\vspace{-10pt}
\label{bg-error}
\end{figure*}

We evaluates the robustness of our proposed system in life-long deployment by examining its performance under environmental variations. 

As illustrated in Fig. \ref{introduction}, the environment represented by the prior map naturally evolves due to factors such as parked vehicles and growing trees. Our system is capable of dynamically updating the map to reflect these changes and can also extend the map to previously unmapped areas. Importantly, our map update strategy is designed to minimize the inclusion of dynamic objects, such as passing vehicles and pedestrians, which helps reduce their negative impact on both mapping and localization accuracy.

Table \ref{APE} presents the Absolute Pose Error (APE) results for each system. We evaluate the error using root-mean-square error (RMSE) and maximum error, calculated from the trajectory results of each system against the ground-turth using the EVO package \cite{evo}. As shown in Table \ref{APE}, with the guidance of prior knowledge, map-based localization methods demonstrate higher localization accuracy compared to SLAM systems that operate without prior map. However, map-based localization methods still have limitations. For example, HDL-Loc fails when the system encounters areas without prior map due to its inability to update maps. 
LiLoc, although capable of updating maps, struggles in dynamic environments such as parking lots with rapidly changing conditions, owing to the delay in map updates. 
In contrast, our system effectively handles both scenarios by dynamically loading local maps and applying the three-layer map management approach. Additionally, even in the presence of dynamic objects and significant platform vibrations, which could mislead localization and degrade performance, our system remains minimally affected and maintains robust localization due to its adaptive map updating strategy. \par
And as Fig. \ref{bg-traj} and Fig. \ref{bg-error} show, our system effectively mitigates the impact of platform vibrations on motion estimation and maintains robust and high-precision localization even in a botanical garden with numerous unstructured trees.

\subsection{Exploring Unknown Environments / Incomplete Map}

\begin{figure*}[!t]
\centering
\includegraphics[width=\linewidth]{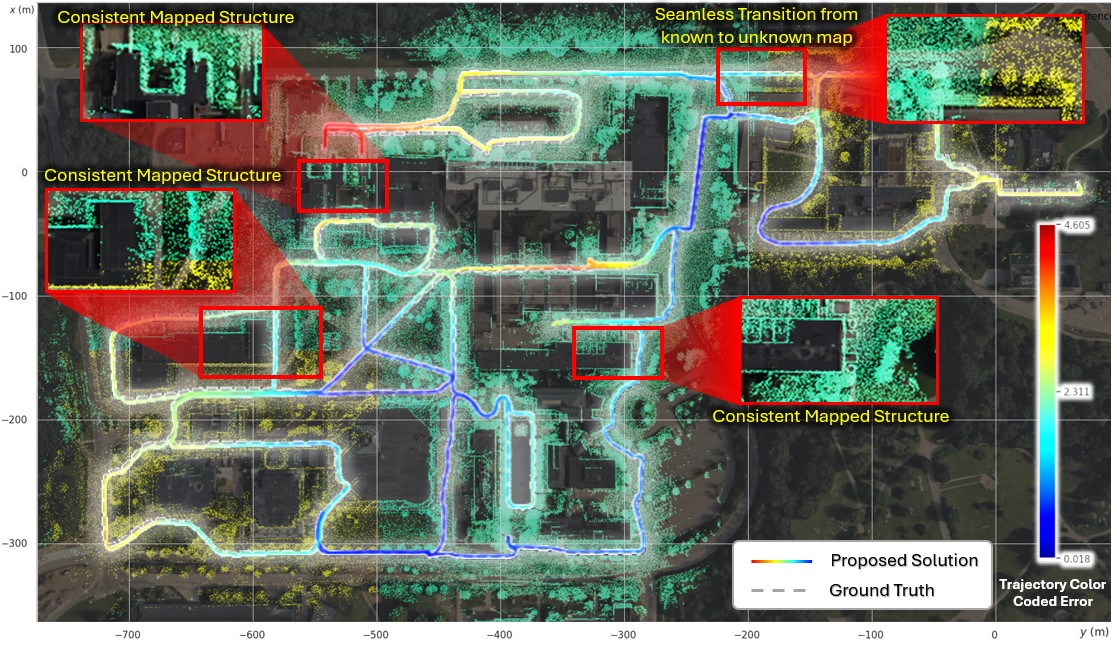}
\vspace{-25pt}
\caption{The constructed map (green points) by LL-Localizer (our proposal) and the prior map (yellow points) on nclt\_4 sequence. The constructed map is compared with the corresponding satellite image. The colorful line shows the trajectory and errors of our proposal, and the dashed line represents the ground-truth trajectory.}
\vspace{-10pt}
\label{long-term}
\end{figure*}

This section further verifies the accuracy and robustness of the proposed system in various environments. \par
To simulate the scenario that the robot traverses from the pre-mapped area to the non-mapped area, stays there for a long period, and then returns to the pre-mapped area, we cropped most of the map of nclt\_4 and conducted the experiment within this modified map. During the operation in the non-mapped area, factors such as random pedestrians, dynamic vehicles, and unstructured trees affect the accuracy of localization. Facing these challenges, the trajectory of the robot may accumulate drift after a long period of movement. \par
In this experiment, Hdl-Loc \cite{HDL-Loc} still fails after entering an area without prior map due to its inability to switch from map-based localization mode to SLAM mode. Fast-LIO-Loc, LTA-OM \cite{LTAOM} and LiLoc \cite{LiLoc} can switch to SLAM mode for map updates; however, their update strategy simply adds scanned point cloud to the map. As a result, even with downsampling process, they still retain many ghost point clouds caused by dynamic objects in the map. As for the SLAM systems Fast-LIO2 \cite{fast-lio2} and Fast-LIO-SLAM, their localization accuracy in mapped areas is inferior to map-based localization methods due to the lack of prior knowledge. In particular, Fast-LIO-SLAM's Scan Context-based loop closure occasionally fails, leading to severe localization errors. But from Fig. \ref{long-term}, it can be seen that LL-Localizer maintains consistently high and excellent localization accuracy in both pre-mapped and non-mapped areas, and achieves updating consistent map after a long period of movement.

\subsection{Time Cost Evaluation}
We conduct experiments using three different methods: managing points within each voxel block using i-Octree, using static Octree, and not using any data structure for point management. After finding the corresponding voxel block via hash function, these three methods perform radius neighbor search using i-Octree, static Octree and traversal, respectively. \par
As Table \ref{Time} shows, compared to traversal and static Octree, i-Octree improves the efficiency of system processing, including neighbor searches and updates of voxel blocks. In radius neighbor search, i-Octree and static Octree exhibit similar efficiency, both significantly outperforming traversal. Additionally, due to its incremental update mechanism, i-Octree achieves faster update efficiency compared to Static Octree. By using Dynamic i-Octree, the majority of frames can be processed within 80ms, and even in dense maps, the system can still maintain real-time operation.

\begin{table}[htbp]
\centering
\caption{Time Cost (millisecond) and Proportion of Time $>$ 80ms ($\%$) Comparison}
\label{Time}
\setlength{\tabcolsep}{6pt}
\renewcommand{\arraystretch}{1.3}
\begin{threeparttable}
    \begin{tabular}{ccccccc}
    \hline
    \multirow{2}{*}{Seq.} & \multicolumn{2}{c}{i-Octree} & \multicolumn{2}{c}{traverse} & \multicolumn{2}{c}{static Octree} \\
    \cmidrule(r){2-3} \cmidrule(r){4-5} \cmidrule(r){6-7}
     & Mean & $>$80ms & Mean & $>$80ms & Mean & $>$80ms \\
    \hline
    nclt\_2     & \textbf{58.37} & \textbf{18.96} & 88.61 & 56.45 & 76.02 & 40.07 \\
    nclt\_3     & \textbf{56.38} & \textbf{15.21} & 78.48 & 43.60 & 74.50 & 38.71 \\
    nclt\_4     & \textbf{51.53} & \textbf{10.14} & 70.01 & 30.56 & 67.76 & 27.34  \\
    gate\_2    & \textbf{62.73} & \textbf{21.00} & 95.30 & 70.65 & 85.20 & 67.31 \\
    gate\_3    & \textbf{63.64} & \textbf{27.91} & 79.57 & 47.60 & 73.53 & 43.59 \\
    street\_2   & \textbf{64.23} & \textbf{20.28} & 92.96 & 66.79 & 80.24 & 64.27 \\
    street\_3   & \textbf{68.73} & \textbf{27.85} & 91.91 & 67.54 & 84.51 & 64.85 \\
    bg\_2       & \textbf{48.83} & \textbf{7.88} & 70.25 & 31.61 & 62.49 & 20.91 \\
    bg\_3       & \textbf{38.16} & \textbf{1.61} & 56.59 & 11.63 & 54.42 & 9.79 \\
    \hline
    \end{tabular}
\end{threeparttable}
\end{table}

\section{Conclusion}
This letter presents LL-Localizer, a method for life-long localization utilizing prior maps based on incremental voxel.
The framework we proposed allows robots to maintain robust and accurate localization even when the environment changes or they traverses between mapped and unmapped areas, without the ability of environmental perception. And it can simultaneously update the maps of areas where the environment has changed and areas without maps. In addition, we utilised Dynamic i-Octree, which is an improvement from Dynamic Octree. This data structure is very suitable for our map management method and can maintain accurate and efficient neighbor search in maps of different densities through hash map and i-Octree even in dense maps.
Contrast experiments in public datasets verifies that LL-Localizer can perform stable and accurate localization comparable to state-of-the-art localization systems in different environments. And even if the environment in the prior map changes or the robots traverses between mapped and unmapped areas, our system can still maintain consistently robust and accurate localization. 

\bibliographystyle{IEEEtran}
\bibliography{references}

\end{document}